\documentclass{article}

\usepackage{caption}
\usepackage{PRIMEarxiv}
\usepackage[utf8]{inputenc} 
\usepackage[T1]{fontenc}    
\usepackage{hyperref}       
\usepackage{url}            
\usepackage{booktabs}       
\usepackage{amsfonts}       
\usepackage{nicefrac}       
\usepackage{microtype}      
\usepackage{lipsum}
\usepackage{fancyhdr}       
\usepackage{graphicx}       
\usepackage{amsmath}
\usepackage{xcolor}
\graphicspath{{media/}}     
\usepackage{multirow}
\usepackage{cite}

\usepackage{CJKutf8}
\usepackage{makecell}
\usepackage{ulem}
\usepackage{hyperref}

\title{C$^{3}$Bench:  A Comprehensive Classical Chinese Understanding Benchmark for Large Language Models
}

\author{
  \parbox{\linewidth}{\centering Jiahuan Cao, Yongxin Shi, Dezhi Peng, Yang Liu, Lianwen Jin\thanks{Corresponding Author.}
  }
  \\
  \parbox{\linewidth}{\centering\vspace{2mm}
  South China University of Technology~~\\
  }
 \\
 \parbox{\linewidth}{\centering\vspace{2mm}
 \tt jiahuanc@foxmail.com, eelwjin@scut.edu.cn}
 \\
}

\begin{document}
\maketitle
\begin{abstract}
Classical Chinese Understanding (CCU) holds significant value in preserving and exploration of the outstanding traditional Chinese culture. 
Recently, researchers have attempted to leverage the potential of Large Language Models (LLMs) for CCU by capitalizing on their remarkable comprehension and semantic capabilities. 
However, no comprehensive benchmark is available to assess the CCU capabilities of LLMs.
To fill this gap, this paper introduces C$^{3}$bench, a Comprehensive Classical Chinese understanding benchmark, which comprises 50,000 text pairs for five primary CCU tasks, including classification, retrieval, named entity recognition, punctuation, and translation. 
Furthermore, the data in C$^{3}$bench originates from ten different domains, covering most of the categories in classical Chinese.
Leveraging the proposed C$^{3}$bench, we extensively evaluate the quantitative performance of 15 representative LLMs on all five CCU tasks.
Our results not only establish a public leaderboard of LLMs' CCU capabilities but also gain some findings.
Specifically, existing LLMs are struggle with CCU tasks and still inferior to supervised models.
Additionally, the results indicate that CCU is a task that requires special attention.
We believe this study could provide a standard benchmark, comprehensive baselines, and valuable insights for the future advancement of LLM-based CCU research.
The evaluation pipeline and dataset are available at \url{https://github.com/SCUT-DLVCLab/C3bench}.

\end{abstract}

\section{Introduction}
The 5000-year history of Chinese civilization has engendered numerous precious cultural artifacts, with classical Chinese serving as one of the most crucial carriers for this heritage. 
Consequently, classical Chinese understanding plays a significant role in safeguarding and advancing this rich traditional Chinese culture. 
However, due to its distinctive language structure and vocabulary, comprehending classical Chinese poses a significant challenge for non-experts. 
With the rapid development of deep learning, some researchers tried to employ deep models to accomplish CCU tasks, such as named entity recognition\cite{yu2020bert-ner,han2018chinese-term} and translation\cite{jiang2023C2C,chang2021time-aware}.
Recently, Large Language Models (LLMs) have demonstrated impressive capabilities in various NLP tasks\cite{raffel2020T5,zhang2022opt,chung2022flan-t5,chowdhery2023palm,brown2020gpt3,openai2023gpt4}.

\begin{figure}
    \centering
    \includegraphics[width=10cm,height=8cm]{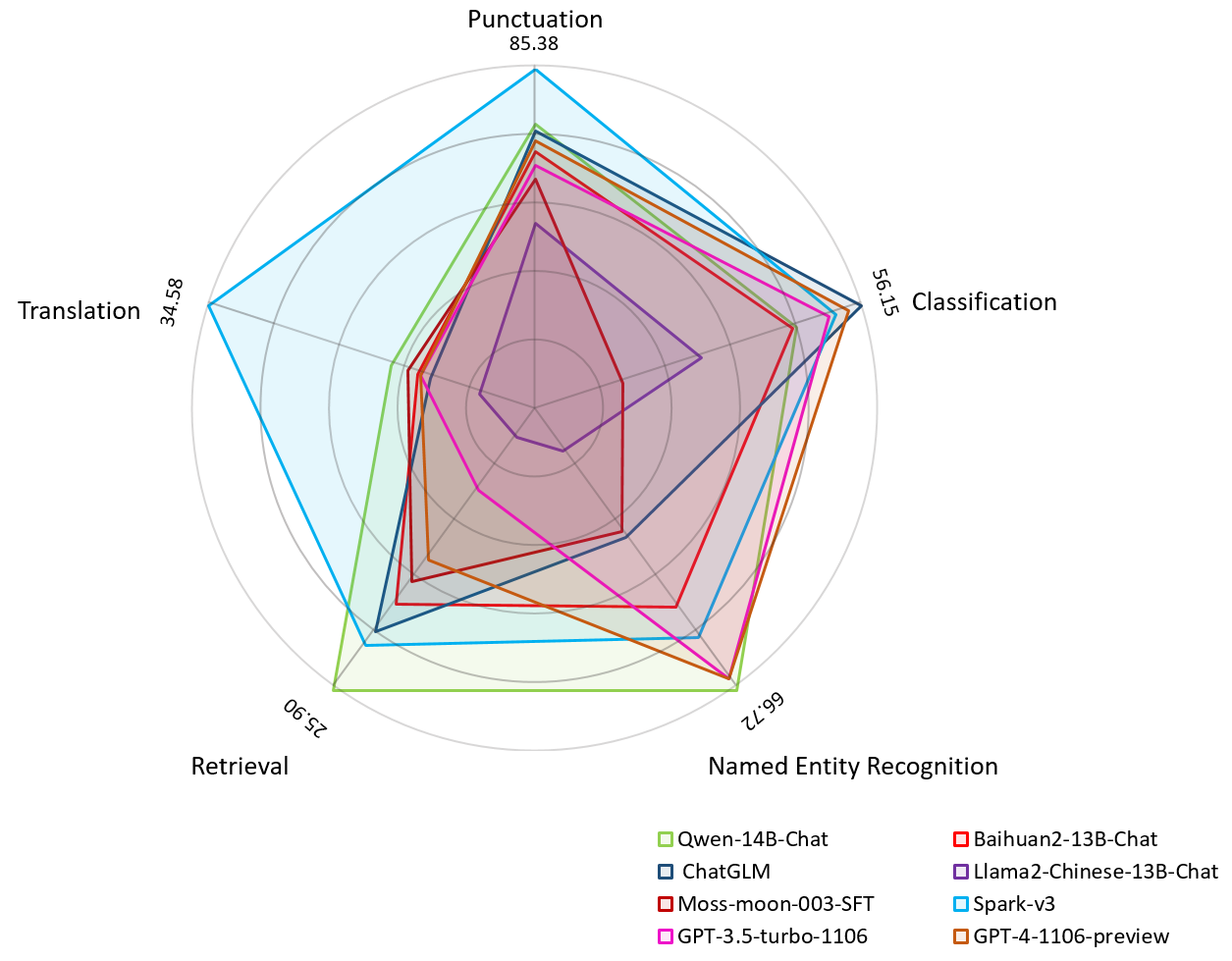}
    \caption{The performance of several representative evaluated models on C$^{3}$Bench.}
    \label{fig:all_rank}
\end{figure}

Benefiting from large-scale training corpus, LLMs possess robust transfer abilities\cite{chi2023low-resource,ghazvininejad2023low-resource,yohannes2022low-resource}. Some researchers have delved into the potential of LLMs for CCU tasks and achieved preliminary success\cite{chang2023sikugpt,cao2023ALT,wang2023evahan2023}.
Nevertheless, the classical Chinese understanding capability of LLMs remains inadequately explored due to the lack of a standard CCU benchmark, hampering further investigation in this field.

To fill this gap, we propose C$^{3}$bench, a \textbf{C}omprehensive \textbf{C}lassical \textbf{C}hinese understanding benchmark for LLMs, which consists of 50,000 text pairs for five CCU tasks, including classification, retrieval, named entity recognition, punctuation, and translation. 
Unlike previous benchmark\cite{ji2021c-clue} for specific CCU tasks or  benchmark\cite{zhang2023aclue} that use simple multiple choice questions to evaluate generative large language models, C$^{3}$bench is the first multi-task CCU benchmark specified for LLMs using natural language generation. 
Furthermore, the C$^{3}$bench consists of data from ten different domains, requiring models to possess a broad and diverse range of understanding capabilities, aligning closely with the inherent attributes of LLMs.
Based on the C$^{3}$bench, we conduct extensive evaluations of 15 representative models, affirming their significant potential for future academic research endeavors. We draw the performance of several representative models in a radar map as shown in Figure~\ref{fig:all_rank}.
The evaluation results unveil that existing LLMs are struggle with CCU tasks and still inferior to supervised models.
Additionally, the results indicate that CCU is a task that requires special attention.
In summary, our contributions are as follows:
\begin{itemize}
    \item We propose a comprehensive classical Chinese understanding benchmark, named C$^{3}$bench. To the best of our knowledge, this is the first classical Chinese understanding benchmark for LLMs using natural language generation.
    \item Based on C$^{3}$bench, we comprehensively evaluate 15 widely-used models, revealing and quantifying their abilities in classical Chinese understanding.
    \item We analyze the evaluation results in depth, uncovering valuable findings that can offer meaningful references and insights for the future advancement of LLM-based CCU research.
\end{itemize}

\section{Related works}
\label{related_works}

\subsection{Large language models}
The rise of Large language models (LLMs) can be attributed to the introduction of the Transformer\cite{vaswani2017transformer} architecture, which dates back to GPT\cite{radford2018gpt-1} and BERT\cite{kenton2019bert}. Subsequently, the parameters of LLMs expanded rapidly. T5\cite{raffel2020T5} unified the form of tasks, enabling a single model to address multiple tasks. Flan-T5\cite{chung2022flan-t5} used instruction tuning to align the model's responses with human instructions, and Reinforcement Learning from Human Feedback (RLHF)\cite{ouyang2022instruct-GPT} further strengthened this process. ChatGPT\cite{ouyang2022instruct-GPT} and GPT-4\cite{openai2023gpt4} are the most representative LLMs, while LLaMA\cite{touvron2023llama} is the most prominent open-source LLMs, driving the rapid development of the open-source community and giving birth to models like Vicuna\cite{peng2023vicuna}. For Chinese LLMs, GLM-130B\cite{zeng2022glm130b} is a large model based on the General Language Model (GLM) architecture\cite{du2022glm}. Baichuan2\cite{baichuan2023baichuan2} is a series of LLMs that have been trained from scratch on 2.6T tokens, while Qwen\cite{bai2023qwen} is a series of LLMs trained on 3T tokens. Those models have all undergone the Supervised Fine-Tuning (SFT)\cite{chung2022flan-t5} and RLHF\cite{ouyang2022instruct-GPT} processes to get the chat models.

\subsection{Automatically classical Chinese understanding}
Early automatic systems for classical Chinese comprehension originated from the application of traditional Natural Language Processing (NLP) techniques. These models were tailored for specific tasks, such as punctuation, named entity recognition, and translation. 
These approaches\cite{zhang2015character-words, jiang2023C2C} predominantly rely on intricate statistical models or extensively hand-annotated supervised data to attain satisfactory performance. Although these methodologies excel in specific tasks, they are often hampered by the constraints of their manually-engineered features and proprietary resources, which inhibit their capacity to generalize across a spectrum of tasks.
With the rise of deep learning in NLP, we have witnessed the proliferation of large-scale models and transfer learning techniques across a myriad of tasks. Models like GuwenBERT\protect\footnotemark[1] and SikuBERT\cite{wang2023gujibert} exemplify this trend, leveraging pre-trained BERT\cite{kenton2019bert} embeddings specific to classical Chinese. With minimal supervised data, they have achieved remarkable proficiency in classical Chinese understanding. A salient feature of these approaches is their reduced dependency on dictionaries and intricate structures, offering a more generalized means of Classical Chinese understanding.
The emergence of models like SikuGPT\cite{chang2023sikugpt} underscores the benefits of pre-trained models combined with extensive classical literature corpora, especially in generating classical prose and poetry. 
Models such as Bloom-7B-Chunhua\protect\footnotemark[5], utilizing auto-generated fine-tuning data coupled with open-source base models, have achieved high levels in classical Chinese question-answering tasks.

\subsection{Chinese NLP evaluation benchmarks}
Evaluation benchmarks for Natural Language Processing (NLP) in Chinese are crucial for the rapid development of this field.
CLUE\cite{xu2020clue} is widely recognized as one of the most authoritative benchmarks in the field. The introduction of CUGE\cite{yao2021cuge} has further enhanced the assessment of generative capacity.
SuperCLUE\cite{xu2020clue} offers a comprehensive assessment of LLMs from various perspectives, including code, computation, and conversation.
C-Eval\cite{huang2023c-val} compromises evaluations across 52 disciplines within the humanities and social sciences.
CMMLU\cite{li2023cmmlu} covers 67 topics ranging from basic disciplines to advanced professional levels.
AGIEval\cite{zhong2023agieval} is derived from 20 official, public, and high-standard admission and qualification exams intended for general human test-takers.
PromptBench\cite{zhu2023promptbench} and HalluQA\cite{cheng2023HalluQA} assess LLMs concerning robustness and susceptibility to hallucination. 
However, there is no benchmark in the field for assessing LLMs' comprehension of classical Chinese.

\subsection{Classical Chinese benchmarks}
CASIA-AHCDB\cite{xu2019casia} provides more than 2.2 million texts from 10,350 categories for character recognition. 
MTHv2\cite{ma2020mthv2} consists of 3,199 images of Buddhist texts, which can be used for text detection, text recognition, and reading order prediction.
ICDAR 2019 HDRC-CHINESE\cite{saini2019icdar} is a large-scale family records dataset, which contains 12,850 images.
SCUT-CAB\cite{cheng2022scutcab} is a complex dataset for layout analysis. 
M$^{5}$HisDoc~\cite{shi2023m5hisdoc} is a comprehensive Chinese historical document analysis benchmark, which features a wide range of styles.
However, these datasets are specifically designed for visual tasks. 
Regarding the NLP benchmarks of classical Chinese, the C-CLUE\cite{ji2021c-clue} has developed a dataset for named entity recognition and relation extraction. NiuTrans\protect\footnotemark[2] provides an extensive corpus of Classical-Modern Chinese bilingual data. 
In summary, compared with visual benchmarks, there is still a lack of NLP benchmarks of classical Chinese, especially those designed for LLMs. This is one of the reasons that we propose the C$^{3}$Bench dataset, a multitask benchmark tailored for the evaluation of LLMs in classical Chinese.

\footnotetext[1]{https://github.com/ethan-yt/guwenbert}
\footnotetext[2]{https://github.com/NiuTrans/Classical-Modern}

\section{C$^{3}$Bench}
\label{benchmark}
\subsection{Task definition}
To comprehensively evaluate the classical Chinese understanding capabilities of LLMs, we have integrated the following five tasks into the proposed C$^{3}$bench. We provide some examples of each task in Table~\ref{tab:examples}.
\begin{itemize}
    \item\textbf{Classification:} The model is required to assign a given classical Chinese sentence to one of ten categories, namely poetry, history, Buddhism, Confucianism, Taoism, medicine, arts, military, law, and agriculture.
    \item\textbf{Retrieval:} This task requires the model to accurately identify and return the title of the article from which the classical Chinese sentence originated.
    \item\textbf{Named Entity Recognition:} Given a classical Chinese sentence, this task requires the model to return the named entities (for example, the personal names, official position, or geographical names) within it.   
    \item\textbf{Punctuation:} The model is supposed to insert appropriate punctuation marks into an unpunctuated sentence.
    \item\textbf{Translation:} This task requires the model to translate a classical Chinese sentence into modern Chinese. 
\end{itemize}

\begin{table*}[h]
\centering
\caption{Examples of input and output in C$^{3}$Bench.}
\resizebox{1.\columnwidth}{!}{
\begin{tabular}{lcc}
\toprule
Task & Input & Output\\
\midrule
\vspace{6pt}
Classification & \begin{CJK*}{UTF8}{gbsn}\makecell{天生我材必有用，千金散尽还复来。}\end{CJK*} & \begin{CJK*}{UTF8}{gbsn}{诗}\end{CJK*} \\
\vspace{6pt}
Retrieval & \begin{CJK*}{UTF8}{gbsn}\makecell{中通外直，不蔓不枝，\\香远益清，亭亭净植，可远观而不可亵玩焉。}\end{CJK*} & \begin{CJK*}{UTF8}{gbsn}{《爱莲说》}\end{CJK*} \\
\vspace{6pt}
NER & \begin{CJK*}{UTF8}{gbsn}\makecell{泰至渭南，集诸州兵来会。\\诸将以众寡不敌，请且待欢更西以观之。}\end{CJK*} & \begin{CJK*}{UTF8}{gbsn}{泰、渭、欢}\end{CJK*} \\
\vspace{6pt}
Punctuation & \begin{CJK*}{UTF8}{gbsn}\makecell{夫未战而庙算胜者得算多也未战而庙算不胜者得算少也}\end{CJK*} & \begin{CJK*}{UTF8}{gbsn}\makecell{夫未战而庙算胜者，得算多也；\\未战而庙算不胜者，得算少也。}\end{CJK*} \\
Translation & \begin{CJK*}{UTF8}{gbsn}\makecell{孔子既祥，五日弹琴而不成声，十日而成笙歌。}\end{CJK*} & \begin{CJK*}{UTF8}{gbsn}\makecell{孔子在大祥后五天开始弹琴，但弹不成声调；\\在大祥后逾月的又一旬里欢笙，其声调就和谐了。}\end{CJK*}  \\
\bottomrule
\end{tabular}
}
\label{tab:examples}
\end{table*}

\subsection{Benchmark construction}
Our benchmark construction involved several key procedures.
Firstly, we selected ten representative categories of ancient texts as labels for the classification task. For each category, sentences from ancient texts were collected independently from the Internet, while preserving the titles of their source articles as the Ground Truth (GT) for the retrieval task. In total, 10,000 sentences were collected.
Secondly, the modern Chinese translations of these sentences were obtained through online searches and manual annotation, which serve as the GT for the translation task.
Thirdly, we invited professional annotators to label the entities within the sentences, establishing the GT for the named entity recognition (NER) task.
Fourthly, we removed punctuation marks from each sentence to serve as input for the punctuation task, with the original sentences as the GT.
Finally, we conducted a rigorous and thorough double-check of the aforementioned data to ensure its quality.

\subsection{Data statistics}
\begin{figure}[htbp]
\begin{minipage}[t]{0.5\linewidth}
\centering
\includegraphics[scale=0.45]{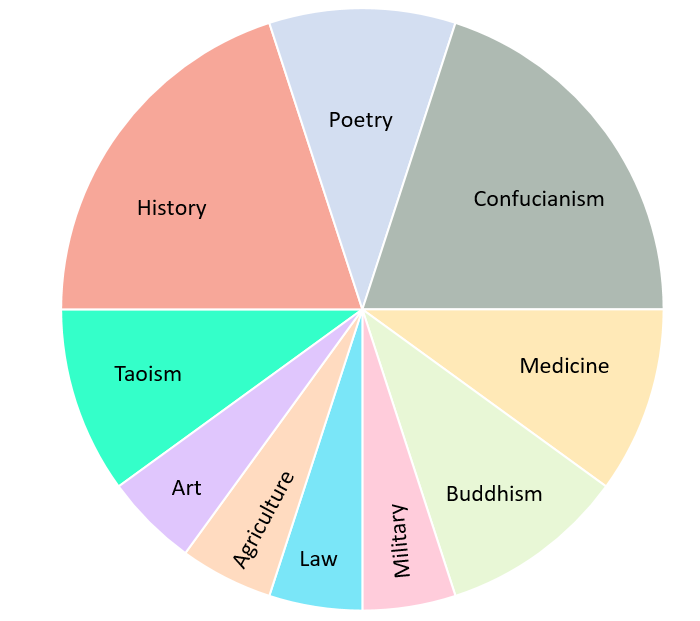}
\caption{Domain distribution of the dataset.}
\label{fig:domain}
\end{minipage}
\begin{minipage}[t]{0.5\linewidth}
\centering
\includegraphics[scale=0.4]{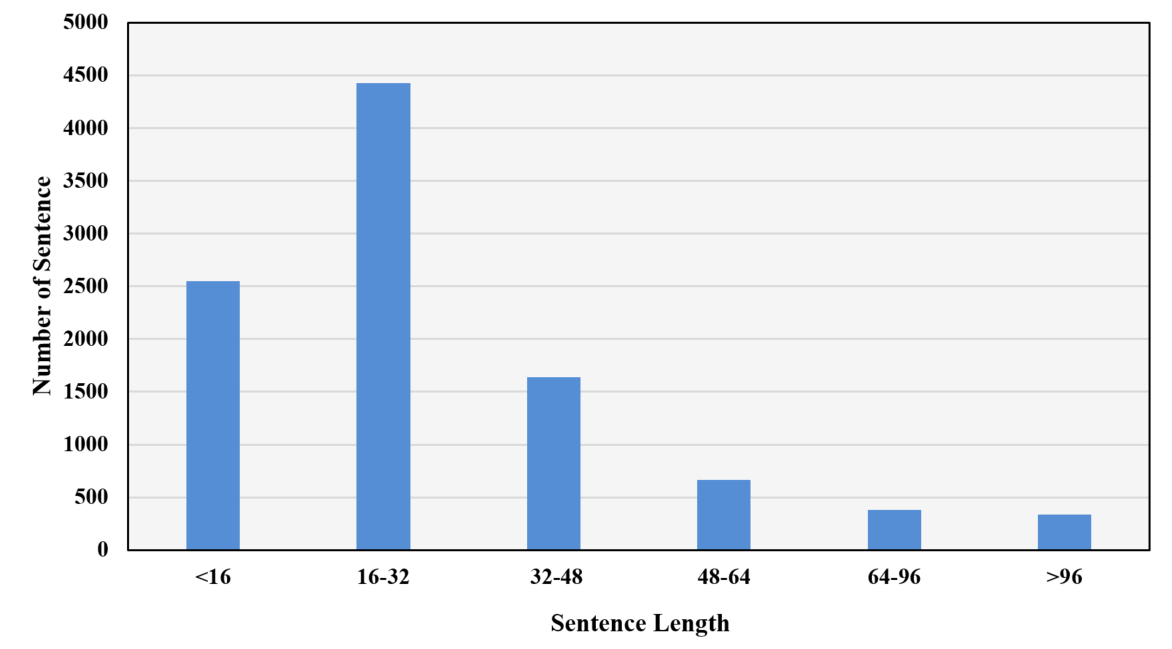}
\caption{Distribution of sentence length.}
\label{fig:length}
\end{minipage}%
\hfill
\end{figure}

The basic data statistics of C$^{3}$bench are shown in Table~\ref{tab:data_statistics}. We can see that each task consists of 10,000 pairs, amounting to a total of 50,000 text pairs. The distribution of domains within the dataset is depicted in Figure~\ref{fig:domain}. History and Confucianism are predominant in classical Chinese texts and constitute the largest proportion of our benchmark. 
As illustrated in Figure~\ref{fig:length}, the coverage of sentence length is quite diverse. Moreover, the majority of sentences have a length ranging from 16 to 32, aligning with the typical distribution in classical Chinese. Additionally, there are also challenging sentences
with length exceeding 96.

\begin{table*}[h]
\centering
\caption{The data statistics of C$^{3}$Bench.}
\begin{tabular}{llcccc}
\toprule
Domain & Classification & Retrieval & Named Entity Recognition & Punctuation & Translation \\
\midrule
Poetry & 1,000 & 1,000 & 1,000 & 1,000 & 1,000 \\
History & 2,000 & 2,000 & 2,000 & 2,000 & 2,000 \\
Buddhism & 1,000 & 1,000 & 1,000 & 1,000 & 1,000 \\
Confucianism & 2,000 & 2,000 & 2,000 & 2,000 & 2,000 \\
Taoism & 1,000 & 1,000 & 1,000 & 1,000 & 1,000 \\
Medicine & 1,000 & 1,000 & 1,000 & 1,000 & 1,000 \\
Art & 500 & 500 & 500 & 500 & 500 \\
Military & 500 & 500 & 500 & 500 & 500 \\
Law & 500 & 500 & 500 & 500 & 500 \\
Agriculture & 500 & 500 & 500 & 500 & 500 \\
 \midrule
Total & 10,000 & 10,000 & 10,000 & 10,000 & 10,000 \\
\bottomrule
\end{tabular}
\label{tab:data_statistics}
\end{table*}

\section{Experiments}
\label{Experiments}

\subsection{Models}
\label{Sec:models}
In this paper, we focus on 15 widely used LLMs, including both open-source models and closed-source models. The detail of these models are presented in Table \ref{tab:detail_models}. We also consider some supervised method for comparison.

\paragraph{Open-source models:} LLaMA\cite{touvron2023llama} is a representative open-source LLM, we selected the modern Chinese localized versions of it, specifically LLaMA2-Chinese-7B-Chat\protect\footnotemark[3] and LLaMA2-Chinese-13B-Chat\protect\footnotemark[4]. We included Bloom-7B-Chuhua\protect\footnotemark[5], which is a classical Chinese localized version of Bloom\cite{scao2022bloom}. Additionally, We selected the most widely recognized ones developed by domestic research institutions for evaluation, including Baichuan2-7B-Chat\cite{baichuan2023baichuan2}, Baichuan2-13B-Chat\cite{baichuan2023baichuan2}, ChatGLM2-6B\cite{du2022glm}, Qwen-7B-Chat\cite{bai2023qwen}, Qwen-14B-Chat\cite{bai2023qwen} and Moss-moon-003-SFT\cite{sun2023moss}.
\paragraph{Closed-source models:} We selected models developed by OpenAI, including GPT-3.5-turbo\cite{ouyang2022instruct-GPT} and GPT-4\cite{openai2023gpt4}, as well as some representative models developed by domestic institutions, including ERNIE-bot-turbo\cite{zhang2019ernie}, Spark-v3\cite{spark}, ChatGLM\cite{du2022glm} and abab5-chat\cite{minimax}.
\paragraph{Supervised methods:} Supervised methods of NER and punctuation are reached by Guwen-NER\protect\footnotemark[6] and Guwen-punc\protect\footnotemark[7], respectively,  both of which are finetuned from GuwenBERT, a RoBERTa pre-trained with 1.7B classical Chinese tokens. Similarly, Guwen-cls\protect\footnotemark[8] is the supervised method of classification, and it is noteworthy that the category of Guwen-cls is different from our category, so only the common category was considered when testing.
And the best result of translation is achieved by Cao, et al\cite{cao2023ALT}, with a LLaMA-13B model finetuned using 400M translation data.

\begin{table*}[h]
\centering
\caption{Details of tested models.}
\resizebox{\textwidth}{!}{
\begin{tabular}{lccccccc}
\toprule
 Model & Access method & Website & Parameters & Release date \\
\midrule
Bloom-7B-Chunhua\protect\footnotemark[5] & Local inf. & https://huggingface.co/wptoux/bloom-7b-chunhua & 7B & April, 2023 \\
Baichuan2-7B-Chat\cite{baichuan2023baichuan2} & Local inf. & https://huggingface.co/baichuan-inc/Baichuan2-7B-Chat & 7B & September, 2023 \\
Baichuan2-13B-Chat\cite{baichuan2023baichuan2} & Local inf. & https://huggingface.co/baichuan-inc/Baichuan2-13B-Chat & 13B & September, 2023 \\
ChatGLM2-6B\cite{du2022glm} & Local inf. & https://huggingface.co/THUDM/chatglm2-6b & 6B & June, 2023 \\
Qwen-7B-Chat\cite{bai2023qwen} & Local inf. & https://huggingface.co/Qwen/Qwen-7B-Chat & 7B & September, 2023 \\
Qwen-14B-Chat\cite{bai2023qwen} & Local inf. & https://huggingface.co/Qwen/Qwen-7B-Chat & 14B & September, 2023 \\
LLaMA2-Chinese-7B-Chat\protect\footnotemark[3] & Local inf. & https://huggingface.co/FlagAlpha/Llama2-Chinese-7b-Chat & 7B & July, 2023 \\
LLaMA2-Chinese-13B-Chat\protect\footnotemark[4] & Local inf.  & https://huggingface.co/FlagAlpha/Llama2-Chinese-13b-Chat & 13B & July, 2023 \\
Moss-moon-003-SFT\cite{sun2023moss} & Local inf. & https://huggingface.co/fnlp/moss-moon-003-sft & 16B & April, 2023 \\
GPT-3.5-turbo\cite{ouyang2022instruct-GPT} & API call & https://chat.openai.com/ & - & November, 2023 \\
GPT-4\cite{ouyang2022instruct-GPT} & API call & https://chat.openai.com/ & - & November, 2023 \\
ERNIE-bot-turbo\cite{zhang2019ernie} & API call & https://console.bce.baidu.com/qianfan & - & September, 2023 \\
Spark-v3\cite{spark} & API call & https://xinghuo.xfyun.cn/sparkapi & - & October, 2023 \\
abab5-chat\cite{minimax} & API call & https://api.minimax.chat/ & - & - \\
ChatGLM$\_$Turbo\cite{du2022glm} & API call & https://open.bigmodel.cn/ & - & October, 2023 \\
\bottomrule
\end{tabular}}
\label{tab:detail_models}
\end{table*}

\subsection{Settings}
\label{Sec:settings}
\paragraph{Prompts} 
The models we chose for evaluation have all undergone instruction-based fine-tuning or Reinforcement Learning from Human Feedback (RLHF)\cite{ouyang2022instruct-GPT}, and thus can follow the instructions of users. 
Consequently, we have constrained the output format of each task in prompts to facilitate automatic and precise evaluation of the output results. The prompts in Chinese for each task are presented in Table \ref{tab:prompts}, with translations provided in English for a wider audience.

\begin{table*}[h]
\centering
\caption{Prompts for each task. We use the Chinese prompt and also provide its English meaning for reference.}
\resizebox{1.\columnwidth}{!}{
\begin{tabular}{lcc}
\toprule
Task & Prompt & Prompt in English\\
\midrule
\vspace{6pt}
Classification & \begin{CJK*}{UTF8}{gbsn}\makecell{不需要解释，\\ 直接从“诗、史、儒、道、佛、农、法、艺、医、兵”中\\选择一个类别输出，\\ 判断下列文言文的类别：[文言文]}\end{CJK*} & \makecell{Without explanation, \\directly from "Poetry, History, Confucianism, Taoism, Buddhism, \\ Buddhism, Agriculture, Law, Arts, Medicine, Military" to select \\ a category output, judge the following categories of classical Chinese: [classical Chinese]} \\
\vspace{6pt}
Retrieval & \begin{CJK*}{UTF8}{gbsn}\makecell{不需要解释，直接给出下列文言文出处书名：[文言文]}\end{CJK*} & \makecell{Without explanation, directly give \\ the following source title of classical Chinese: [classical Chinese]} \\
\vspace{6pt}
NER & \begin{CJK*}{UTF8}{gbsn}\makecell{找出下列文言文中的命名实体。若无实体，输出“无”；\\若有实体，直接输出实体，多个实体之间用“、”分隔：[文言文]}\end{CJK*} & \makecell{Find the following named entities in classical Chinese. \\ If there is no entity, output "no"; if there is an entity, \\ directly output entity, multiple entities are separated by ",": [classical Chinese]} \\
\vspace{6pt}
Punctuation & \begin{CJK*}{UTF8}{gbsn}\makecell{不需要解释，直接输出下列文言文添加标点符号后的结果：[文言文]}\end{CJK*} & \makecell{Without explanation, directly output the following results \\ after adding punctuation marks to classical Chinese: [classical Chinese]} \\
Translation & \begin{CJK*}{UTF8}{gbsn}\makecell{不需要解释，直接输出下列文言文的白话文翻译：[文言文]}\end{CJK*} & \makecell{Without explanation, directly output the \\ following classical Chinese translation of vernacular Chinese: [classical Chinese]}  \\
\bottomrule
\end{tabular}
}
\label{tab:prompts}
\end{table*}

\paragraph{Inference settings}

For open-source models, we employed the recommended system message and role, and loaded the models using bfloat16 precision. Furthermore, the maximum number of tokens for model responses was limited to 2048, with nucleus sampling configured at 1 and top-k sampling at 50. As for closed-source models, we were unable to control the model loading precision, but the other settings remained consistent with those of the open-source models.
It is crucial to note that model outputs may not always strictly conform to the stipulated format. Therefore, post-processing adjustments were implemented to ensure fair and consistent assessments across different models.

\subsection{Metric} 
\label{Sec:Metric}
For different tasks, we employ distinct evaluation metrics. 
\begin{itemize}
    \item \textbf{Classification:} We utilize accuracy in this task. If the model correctly classifies a sentence, it will get a score of 1, or 0 otherwise. The final total score is averaged for the accuracy.
    \item \textbf{Retrieval:} Similarly, we use accuracy for the retrieval task, while due to the multi-level label structure, we calculate accuracy for each level of the hierarchy. 
    \item \textbf{Punctuation:} We adopt F1 score as the evaluation metric as in previous work\cite{zhang2020punc-f1}.
    \item \textbf{Named Entity Recognition:} We utilize F1 score for this task as in previous work\cite{liu2022ner-f1}.
    \item \textbf{Translation:} For this task, we use BLEU\cite{papineni2002bleu} as the evaluation metric.  
\end{itemize}

\subsection{Results and analysis}
The results of all tested models are shown in Table~\ref{tab:res_models}, and we also draw the performance of several representative models in a radar map as shown in Figure~\ref{fig:all_rank}. Based on the results, we draw the following insights:
\label{Sec:results}
\begin{table*}[h]
\centering
\caption{Results of all tasks. The bold and underlined indicate the best and the second best results, respectively.}
\resizebox{\textwidth}{!}{
\begin{tabular}{lcccccc}
\toprule
 Model & Classifications ↑ & Retrieval ↑ & NER ↑ & Punctuation ↑ & Translation ↑ & Average ↑ \\
\midrule
Bloom-7B-Chunhua\protect\footnotemark[5] & 39.62 & 13.36 & 34.70 & 62.19 & 11.27 & 33.23 \\
Baichuan2-7B-Chat\cite{baichuan2023baichuan2} & 37.00 & 18.36 & 63.25 & 53.96 & 13.70 & 37.15 \\
Baichuan2-13B-Chat\cite{baichuan2023baichuan2} & 44.26 & 17.79 & 46.67 & 65.11 & 12.45 & 37.26 \\
ChatGLM2-6B\cite{du2022glm} & 50.28 & 9.03 & 28.56 & 28.48 & 6.76 & 24.62 \\
Qwen-7B-Chat\cite{bai2023qwen} & 49.65 & 13.92 & 28.33 & 69.61 & 15.61 & 35.42 \\
Qwen-14B-Chat\cite{bai2023qwen} & 44.93 & \textbf{25.90} & \uline{66.72} & 71.83 & 15.38 & 44.95 \\
LLaMA2-Chinese-7B-Chat\protect\footnotemark[3] & 18.78 & 3.20 & 12.62 & 34.73 & 4.24 & 14.71 \\
LLaMA2-Chinese-13B-Chat\protect\footnotemark[4] & 28.75 & 2.27 & 9.31 & 47.27 & 5.91 & 18.70 \\
Moss-moon-003-SFT\cite{sun2023moss} & 15.07 & 15.84 & 28.90 & 58.39 & 13.35 & 26.30 \\
GPT-3.5-turbo\cite{ouyang2022instruct-GPT} & 50.65 & 7.36 & 63.83 & 61.34 & 11.94 & 39.02 \\
GPT-4\cite{ouyang2022instruct-GPT} & 53.88 & 13.71 & 63.87 & 67.31 & 12.09 & 42.17 \\
ERNIE-bot-turbo\cite{zhang2019ernie} & 50.70 & 21.22 & 9.61 & 65.29 & 10.66 & 31.50 \\
Spark-v3\cite{spark} & 51.61 & \uline{21.83} & 53.81 & \textbf{85.38} & \uline{34.58} & 49.44 \\
abab5-chat\cite{minimax} & 52.20 & 15.53 & 34.64 & 65.42 & 10.56 & 35.67 \\
ChatGLM$\_$Turbo\cite{du2022glm} & \uline{56.15} & 20.49 & 30.04 & 69.72 & 10.91 & 37.46 \\
\midrule
Supervised-Method & \textbf{82.66}\protect\footnotemark[8] & - & \textbf{73.73}\protect\footnotemark[6] & \uline{82.48}\protect\footnotemark[7] & \textbf{52.02}\cite{cao2023ALT} & - \\
\bottomrule
\end{tabular}}
\label{tab:res_models}
\end{table*}
\footnotetext[3]{https://huggingface.co/FlagAlpha/Llama2-Chinese-7b-Chat}
\footnotetext[4]{https://huggingface.co/FlagAlpha/Llama2-Chinese-13b-Chat}
\footnotetext[5]{https://huggingface.co/wptoux/bloom-7b-chunhua}
\footnotetext[6]{https://huggingface.co/ethanyt/guwen-ner}
\footnotetext[7]{https://huggingface.co/ethanyt/guwen-punc}
\footnotetext[8]{https://huggingface.co/ethanyt/guwen-cls}

\begin{figure}
\centering
\includegraphics[scale=0.7]{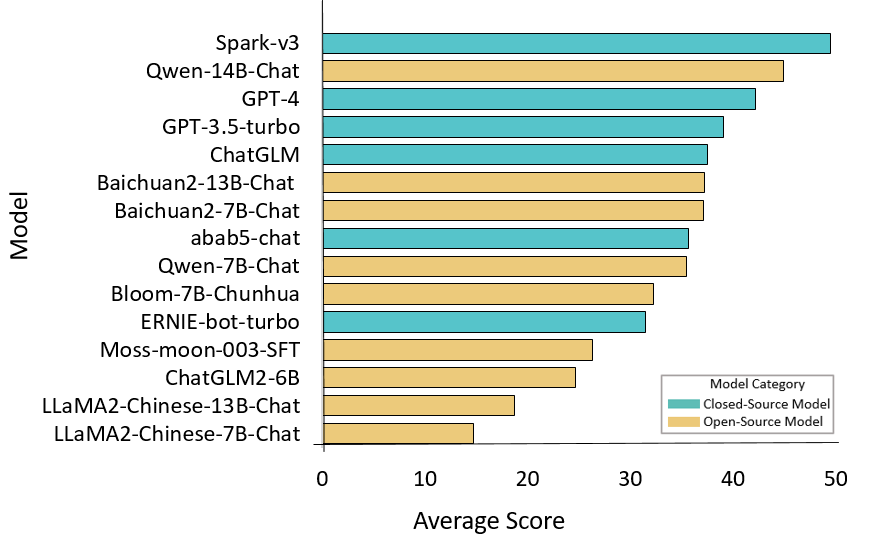}
\caption{Overall ranking of Classical Chinese Understanding capability for all the evaluated models.}
\label{fig:rank}
\hfill
\end{figure}

\paragraph{CCU is challenging for current LLMs:} The overall ranking of the CCU capability, as illustrated in Figure~\ref{fig:rank}, indicates that all 15 tested models scored an average below 50. In the critical translation task in classical Chinese, only one model exceeds a 20 BLEU score. For the highly challenging retrieval task, the accuracy of all models falls below 0.3. This highlights the complexity of our data and suggests significant room for improvement in the ability of LLMs to comprehend classical Chinese.
\paragraph{Existing LLMs do not outperform the supervised models: }
As shown in the last row of Table~\ref{tab:res_models}, the performance of supervised models on specific tasks outperforms existing LLMs, especially demonstrating significant superiority in translation and classification tasks. In classification task, the best supervised model can achieve an accuracy of 0.8266, which is 0.2651 higher than the best LLM. While in translation task, the best LLM achieves only 34.58 BLEU, far behind the supervised method of 52.02 BLEU.

\paragraph{Current LLMs have unbalanced CCU capability:} Classical Chinese understanding capability of some large models is unbalanced.  
For example, ChatGLM$\_$Turbo\cite{du2022glm} ranks first in the classification task and also achieves good retrieval and punctuation performance. However, it exhibits extremely low performance on the named entity recognition and translation tasks.
We argue that this may be due to the imbalanced training data associated with individual tasks.
\paragraph{Classical Chinese understanding is a task that requires special attention:} Although certain studies\cite{cui2023Chinese-llama} have indicated significant improvements in the modern Chinese understanding capabilities of English models such as LLaMA after fine-tuning with a small amount of Chinese data, our experiments suggest that such models have insufficient understanding capabilities of classical Chinese. This underscores the unique requirements for understanding classical Chinese, which demands knowledge of Chinese characteristics and traditional culture not readily transferable from English. Therefore, improving the classical Chinese understanding capability of LLMs is a critical concern that necessitates special attention.

\section{Limitations}

There are two main limitations of our work.
First, the C$^{3}$bench does not exhaustively cover all types of classical Chinese understanding (CCU) tasks, such as the summary of ancient texts. Instead, we focus on five representative tasks, and hope to evaluate the CCU capabilities of LLMs through these tasks.
Second, only the zero-shot capacity in CCU tasks was evaluated, without exploring few-shot scenarios.

\section{Conclusion}

In this study, we introduce C$^{3}$bench, the first comprehensive Classical Chinese Understanding (CCU) benchmark for Large Language Models (LLMs), which spans a multitude of tasks and various domains. 
Based on the C$^{3}$bench, we comprehensively evaluate 15 representative LLMs, quantifying their capabilities in CCU and establishing a public leaderboard. 
Our results unveil that existing LLMs are struggle with CCU tasks and still inferior to supervised models.
Additionally, the results indicate that CCU is a task that requires special attention.
This study provides a standard benchmark, comprehensive baselines, and valuable insights for the future advancement of LLM-based CCU research.


\bibliographystyle{unsrt}  
\bibliography{main}

\end{document}